%% file: main.tex
\documentclass[runningheads]{llncs}
\usepackage[T1]{fontenc}
\usepackage{graphicx}
\input{math_commands.tex}

\input{meta/macro}

\begin{document}
\title{An Attention-based Representation Distillation Baseline for Multi-Label Continual Learning}
\titlerunning{SCAD: a new baseline for Multi-Label Continual Learning}
%
\author{
Martin Menabue\inst{1} \and
Emanuele Frascaroli\inst{1} \and
Matteo Boschini\inst{1} \and
Lorenzo Bonicelli\inst{1} \and
Angelo Porrello\inst{1} \and
Simone Calderara\inst{1}
}

\authorrunning{M. Menabue et al.}
\institute{
$^1$
University of Modena and Reggio Emilia, Italy \\
\email{\{name.surname\}@unimore.it}
}
%
\maketitle              

\begin{abstract}
\input{contents/00_abstract}

\end{abstract}

\section{Introduction}
\label{sec_introduction}
\input{contents/01_introduction}

\section{Related works}
\label{sec_related_works}
\input{contents/02_related}

\section{Method}
\label{sec_method}
\input{contents/03_method}
\section{Experiments}
\label{sec_experiments}
\input{contents/04_experiments}
\section{Model Analysis}
\label{sec_analysis}
\input{contents/05_analysis}
\section{Conclusions}
\label{sec_conclusions}
\input{contents/06_conclusions}

\begin{credits}
\subsubsection{\ackname} This paper has been supported from Italian Ministerial grant PRIN 2020 ``LEGO.AI: LEarning the Geometry of knOwledge in AI systems'', n. 2020TA3K9N. We acknowledge the CINECA award under the ISCRA initiative, for the availability of high performance computing resources and support.
\subsubsection{\discintname}
The authors have no competing interests to declare that are relevant to the content of this article.
\end{credits}

\bibliographystyle{splncs04}
\bibliography{main}

%
%
%





\end{document}

%% file: math_commands.tex
\usepackage{amsmath,amsfonts,bm}

\def\eqref#1{equation~\ref{#1}}

\def\1{\bm{1}}

\DeclareMathAlphabet{\mathsfit}{\encodingdefault}{\sfdefault}{m}{sl}
\SetMathAlphabet{\mathsfit}{bold}{\encodingdefault}{\sfdefault}{bx}{n}



%% file: meta/macro.tex
\usepackage{xspace}
\usepackage{enumitem}
\usepackage{amsmath}
\usepackage{amssymb}
\usepackage{graphicx}
\usepackage{wrapfig}
\usepackage{makecell}
\usepackage{pifont}
\usepackage{float}
\usepackage{algpseudocode}
\usepackage{algorithm}
\usepackage{setspace}
\usepackage{booktabs}

\newcommand*{\codefont}{\fontfamily{pcr}\selectfont}

\newcommand{\result}[2]{\ensuremath{#1}~\scriptsize{(\ensuremath{#2}})}

\newcommand{\methodname}{Selective Class Attention Distillation\xspace} 
\newcommand{\methnam}{SCAD\xspace} 

\newcommand{\cmark}{\ding{51}}%
\newcommand{\xmark}{\ding{55}}%

%% file: contents/00_abstract.tex
The field of Continual Learning (CL) has inspired numerous researchers over the years, leading to increasingly advanced countermeasures to the issue of \textit{catastrophic forgetting}. Most studies have focused on the single-class scenario, where each example comes with a single label. The recent literature has successfully tackled such a setting, with impressive results. Differently, we shift our attention to the multi-label scenario, as we feel it to be more representative of real-world open problems. In our work, we show that existing state-of-the-art CL methods fail to achieve satisfactory performance, thus questioning the real advance claimed in recent years. Therefore, we assess both old-style and novel strategies and propose, on top of them, an approach called Selective Class Attention Distillation (SCAD). It relies on a knowledge transfer technique that seeks to align the representations of the student network -- which trains continuously and is subject to forgetting -- with the teacher ones, which is pretrained and kept frozen. Importantly, our method is able to selectively transfer the relevant information from the teacher to the student, thereby preventing irrelevant information from harming the student's performance during online training. To demonstrate the merits of our approach, we conduct experiments on two different multi-label datasets, showing that our method outperforms the current state-of-the-art Continual Learning methods. Our findings highlight the importance of addressing the unique challenges posed by multi-label environments in the field of Continual Learning. The code of SCAD is available at {\codefont\url{https://github.com/aimagelab/SCAD-LOD-2024}}.

%% file: contents/01_introduction.tex
During their lifetime, individuals exist within an environment that undergoes continuous changes and are constantly confronted with new and diverse problems. While humans possess the ability to adapt to these changes, acquire new knowledge, and remember past information, studies have shown that artificial intelligence algorithms struggle to replicate this behavior. Specifically, it has been observed that neural networks tend to forget previously solved tasks while learning new knowledge, a phenomenon known as \textit{catastrophic forgetting}~\cite{mccloskey1989catastrophic}. The field of Continual Learning (CL)~\cite{de2021continual,parisi2019continual} has emerged to explore techniques that enable neural networks to continuously learn new concepts while retaining previously acquired knowledge. 

The problem of forgetting has been investigated from various perspectives, leading to the development of numerous effective techniques~\cite{buzzega2020dark,wang2022learning,smith2022coda}.
Many previous works have adopted ResNet-like~\cite{he2016deep} models as a backbone~\cite{buzzega2020dark,rebuffi2017icarl,lopez2017gradient}. However, due to advancements in many Deep Learning fields with the introduction of Transformers~\cite{vaswani2017attention} architectures, recent works have shifted towards using pretrained Vision Transformers (ViT)~\cite{wang2022learning,wang2022dualprompt,smith2022coda}, achieving extraordinary performance. Remarkably, these methods can display results close to the upper bound, \textit{i.e.}, a model trained offline with access to the whole dataset. Thus, we believe current CL image classification benchmarks are close to being saturated, as these new models show almost zero forgetting. This leads us to ask: has the catastrophic forgetting problem been successfully solved?

Present CL settings focus on single-label datasets, where each example comes with a single label. However, these settings may represent an approximation of real-world contexts: in life, individuals are used to observing objects that are simultaneously related to multiple concepts (e.g. a person can be related to a precise age, gender and nationality). To the best of our knowledge, only a few research works have tackled multi-label settings in an incremental fashion~\cite{abdelsalam2021iirc,kim2020imbalanced}.

In this work, we find that modern state-of-the-art CL approaches, that strongly contrast forgetting in single-label scenarios, struggle to deal with multiple labels. In light of these findings, we explore \textbf{Multi-Label Continual Learning} (\textbf{MLCL}) as a new context in which catastrophic forgetting still needs to be adequately addressed. In summary:
\begin{itemize}[leftmargin=*]
    \item We employ two MLCL benchmarks: IIRC CIFAR-100, proposed by~\cite{abdelsalam2021iirc}, which adds superclass prediction to the popular CIFAR-100 dataset~\cite{krizhevsky2009learning}, and Incremental WebVision, a novel real-life scenario extracted from the large WebVision dataset~\cite{li2017webvision}.
    \item We evaluate current SOTA methods based on prompting in the MLCL setting, showing that they fail under the new challenges it poses.
    \item We implemented and tested popular rehearsal CL baselines on top of the pretrained ViT backbone, revealing that these simple combinations are not overwhelmed by MLCL adversities.
    \item We present a new framework able to cope with MLCL difficulties: \textbf{\methodname} (\textbf{\methnam}); inspired by~\cite{ramasesh2022effect,boschini2022transfer}, which highlights the usefulness of pretraining in mitigating the problem of forgetting, our approach employs attention to select valuable information from a pretrained frozen teacher model and transmits it to the student network facing incremental tasks.
\end{itemize}

We show that \methnam achieves the best performance in both MLCL benchmarks, demonstrating that filtering only useful information from pretraining effectively mitigates forgetting in this new challenging scenario.

%% file: contents/02_related.tex
\textbf{Continual Learning} (\textbf{CL})~\cite{de2021continual,parisi2019continual} is a field of Deep Learning that addresses the issue of \textit{catastrophic forgetting}~\cite{mccloskey1989catastrophic}, a phenomenon that occurs when a model is trained on a sequence of tasks. As the data distribution shifts between tasks, the model needs to adapt to new tasks (\textit{plasticity}) while preserving past information (\textit{stability}). 
It would be ideal to archive a balance between these two properties, as too much plasticity may lead the model to forget previously learned information, while excessive stability may prevent the model from learning new tasks.

Continual Learning (CL) approaches are usually divided into the following categories \cite{van2019three}:
\begin{itemize}[leftmargin=*]
    \item \textbf{\textit{Rehearsal based methods}} contrast catastrophic forgetting by storing a subset of samples from past tasks in a memory buffer; these are then replayed during training along with data from the current task. \textbf{Experience Replay}~\cite{ratcliff1990connectionist,robins1995catastrophic} (\textbf{ER}) is a simple baseline that interleaves samples from the buffer with samples from the current task. More advanced methods have been developed from this base technique: \textbf{Dark Experience Replay} (\textbf{DER})~\cite{buzzega2020dark}, which stores the logits of the model to distill knowledge from previous tasks; \textbf{Gradient Episodic Memory} (\textbf{GEM}) \cite{lopez2017gradient} and its improved version \textbf{A-GEM}~\cite{chaudhry2018efficient}, that use past training data to minimize gradient interference.
    \item \textbf{\textit{Regularization based methods}} introduce additional terms in the loss function to prevent the weights of the model from excessively changing; \textbf{Elastic Weight Consolidation} (\textbf{EWC}) \cite{kirkpatrick2017overcoming} estimates the importance of each weight and dynamically constrains their shifts.
    \item \textbf{\textit{Knowledge distillation methods}} \cite{hinton2015distilling} aim to reduce the disparities between a model's current representations and its previous projections. For example, \textbf{Learning Without Forgetting} (\textbf{LWF}) \cite{li2017learning} computes the responses of the model for new examples at the beginning of each task and uses them to guide the training of a student.
\end{itemize}

In recent years, with the advent of Vision Transformers~\cite{vit}, researchers have been exploring Continual Learning solutions using these new architectures. One category of successful methods is represented by the \textbf{\textit{prompting}} approach \cite{wang2022learning,wang2022dualprompt,smith2022coda}, which allocates a set of trainable tokens, called \textit{prompts}, at the beginning of training. These prompts are concatenated with the input of the Vision Transformer and updated throughout the training while the rest of the model typically remains frozen. Researchers have devised various mechanisms for prompt selection to ensure that prompts can encode information about the current task during training. This proves useful during inference as prompts can assist the model in remembering past tasks.

\textbf{Multi-label classification}. Traditional Continual Learning approaches are designed for single-label settings, where each sample is associated with only one label. In multi-label setups~\cite{sorower2010literature,ganda2018survey,TAREKEGN2021107965}, a sample can have multiple labels, and the model must predict all of them accurately.

An interesting multi-label setup is presented in \cite{abdelsalam2021iirc}: the authors propose a novel benchmark called \textbf{IIRC}, which involves adapting widely used datasets, such as CIFAR-100 and ImageNet, for multi-label continual evaluation. This adaptation is achieved by assigning to each sample both a superclass and a subclass label. The authors identified two possible settings: \textit{complete information}, where all labels for each sample are available, and \textit{incomplete information}, where only the label of the current task is provided. We have chosen IIRC as one of our benchmarks and decided to concentrate on the incomplete setup, which the authors denote as a more challenging scenario.

Regarding MLCL literature, an interesting approach is presented in~\cite{kim2020imbalanced}, where the authors aim to tackle two challenges simultaneously: the problem of catastrophic forgetting and the issue of long-tail distribution, in which a few categories have a vast number of samples while most have only a few. The authors' experiments reveal that the least frequent classes (tail) tend to suffer more from forgetting compared to the most recurrent classes (head). To address this issue, they devised a novel sampling strategy called \textit{Partitioned Reservoir Sampling} to select which samples to store in the buffer.

%% file: contents/03_method.tex
\subsection{Multi-Label Continual Learning}
In a multi-label setting, an image classification dataset $\mathcal{D}$ consists of pairs $(\mathbf{x}_i, \mathbf{y}_i)_{i=1}^{|\mathcal{D}|}$, where $\mathbf{x}_i \in \mathbb{R}^{3\times H \times W}$ is an RGB image and $\mathbf{y}_i \in \mathbb{R}^{|\mathcal{Y}|}$ is the multi-hot label vector. Specifically, each element $y_{i,j}$ is 1 if the sample belongs to the class with index $j$, and 0 otherwise. The set $\mathcal{Y}$ encompasses all the possible labels. Moreover, let $\mathcal{B}=\{\mathbf{x}_i, y_i\}_{i=0}^B$ be a $B$-dimensional batch of samples, we denote with $\mathbf{X}[s:e]$ the subset $\{\mathbf{x}_i, y_i\}_{i=s}^e\subseteq\mathcal{B}$.

In the MLCL scenario, a model $f_{\theta}$ is trained on a sequence of $N$ tasks, where each task $T_t = \{(x_j^t,y_j^t)\}_{j=1}^{|T_t|}, t=1,...,N$. An objective function $\mathcal{L}$ is optimized in order to minimize the classification error:
\begin{equation}
    \min_{\theta}{\mathcal{L}_{clf}}=\sum_{t=1}^{N} \left[ \mathbb{E}_{(\mathbf{x},\mathbf{y}) \sim T_t} \left[\ell(\mathbf{y},f_{\theta}(\mathbf{x}))\right] \right]
\label{eq:L_clf}
\end{equation}
where $\ell$ is a loss function, commonly the binary cross-entropy loss \cite{rajeswar2022multi,abdelsalam2021iirc}. Directly optimizing this function is not feasible since we assume that samples from previous tasks are no longer accessible. Consequently, our goal is to find parameters that fit the current task while retaining knowledge from previous tasks.
\subsection{Pretraining and forgetting}
\begin{figure}[h]
\centering
\includegraphics[width=\linewidth]{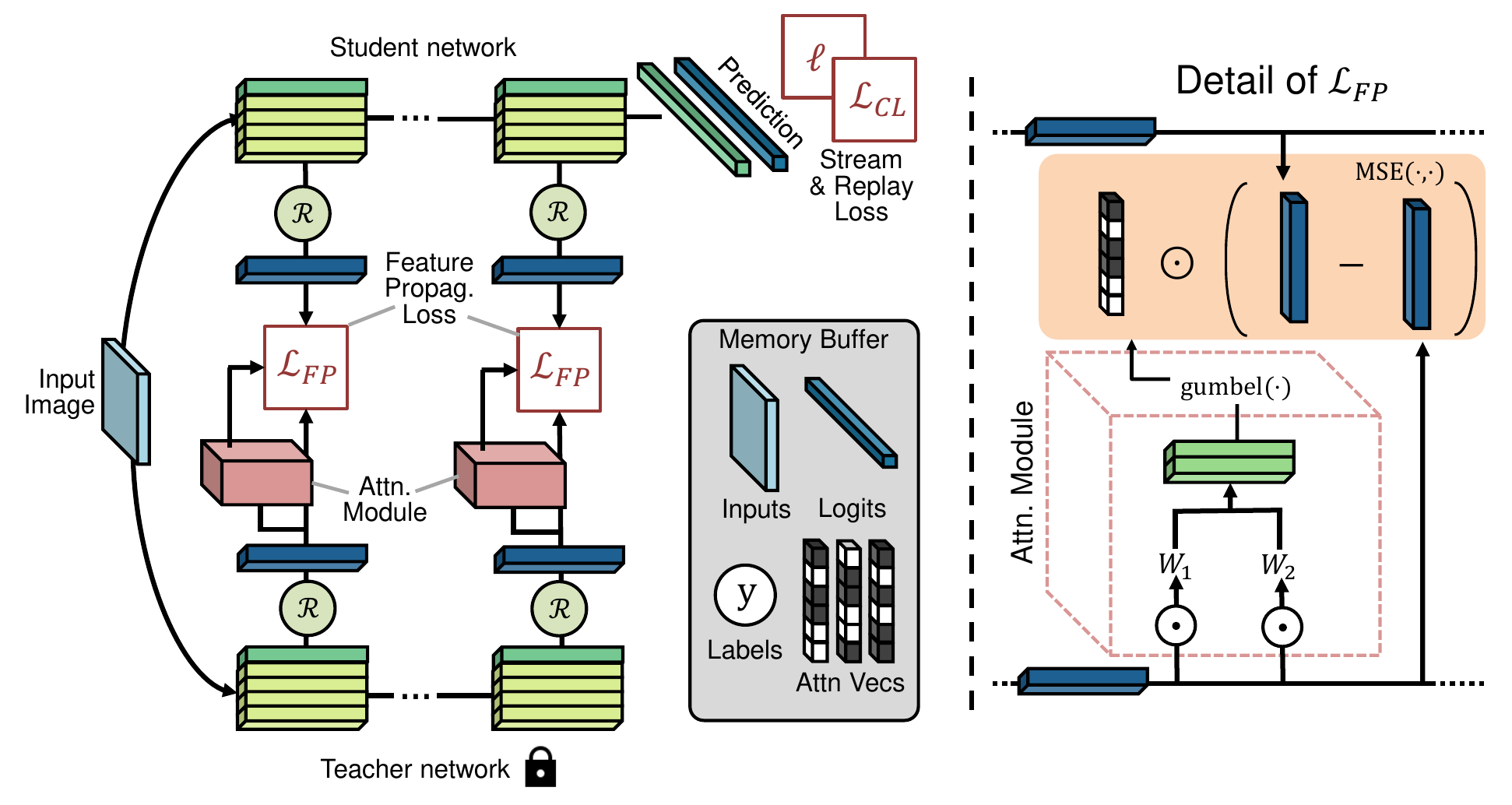}
\caption{\small Our proposal involves two pretrained Vision Transformers: one is frozen, referred to as the \textit{teacher}, while the other is not, referred to as the \textit{student}. An image is provided as input to both the teacher and the student, and intermediate representations are extracted. From these representations, using equation (\ref{eq:attn}) and the indexing indicated in equation (\ref{eq:dist}), we can derive the attention vector that contains the relationships between the class token and the other tokens. These operations are summarized in the image by the operator $\mathcal{R}$. The attention vector is used by the attention modules (\textit{adapters}) to derive binary vectors useful for filtering knowledge transfer. The feature propagation loss $\mathcal{L}_{FP}$ enables maintaining alignment between the attention vectors of the teacher and the student.}
\label{fig:my_label}
\end{figure}
Earlier studies \cite{ramasesh2022effect,mehta2021empirical} have emphasized the resilience of large-scale pretrained models against the issue of catastrophic forgetting. However, as mentioned in \cite{boschini2022transfer}, when a model is trained on a sequence of tasks, it tends to gradually deviate from its initial pretraining state, thereby intensifying the problem of forgetting. To address this issue, we explored solutions to keep a network aligned with the pretraining state during the incremental learning. Our proposal draws inspiration from various research works addressing the problem of knowledge transfer between neural networks~\cite{9747484,boschini2022transfer,lin2022knowledge,yang2022vitkd}. The method we have developed, SCAD, involves using two pretrained neural networks, with one network (referred to as the \textit{teacher}) being frozen, while the parameters of the other network (referred to as the \textit{student}) are optimized during training. In the following paragraphs, we provide a more detailed description of our approach.
\subsection{Selective Class Attention Distillation}
\paragraph{Backbone} Regarding the network architecture we use a pretrained Vision Transformer~\cite{vit} for both the teacher and the student since recent Continual Learning state-of-the-art methods use this kind of architecture. In a Vision Transformer, the input image $\mathbf{x} \in \mathbb{R}^{3 \times H \times W}$ is divided into non-overlapping $S$ patches, which are then flattened and projected by the patch embedding layer, obtaining a sequence of tokens. An additional token, the \textit{class token}, which serves for classification, is concatenated to the other tokens, obtaining a sequence $\mathbf{x}_p \in \mathbb{R}^{(S+1) \times D}$, where $S$ is the sequence length and $D$ is the embedding dimension. The sequence of tokens is forwarded through a chain of transformer encoder blocks, composed by Multi Head Self Attention \cite{vaswani2017attention} and Feed Forward layers, and the final class token is used to make the prediction. Thanks to the Multi-head Self Attention, the model is able to capture different types of dependencies and relationships between image tokens. Moreover, each attention head focuses on different aspects of the input, enabling the model to capture local and global dependencies.

\paragraph{Knowledge transfer technique}
While minimizing the distance between the intermediate representations of teacher and student is a straightforward approach to knowledge transfer, it is not the sole technique available. Recent years have witnessed extensive research aimed at investigating efficient strategies for distilling knowledge among Vision Transformers \cite{chen2022dearkd,yang2022vitkd,9747484}. Our research work draws inspiration from the technique introduced by \cite{9747484}. When passing an input through a Vision Transformer, we can obtain the intermediate representations (i.e. the output of each block), denoted as $F^{l} \in \mathbb{R}^{(S+1) \times D}$, where $l \in 1,...,|L|$ represents the index of the transformer block under consideration. At this stage, we compute the correlation between the tokens within these representations, after normalizing the feature map:
\begin{equation}
\mathcal{R}^l=\frac{F^l}{||F^l||_2^2} \cdot \left(\frac{F^l}{||F^l||_2^2}\right)^T
\label{eq:attn}
\end{equation}
Here, $\mathcal{R}^l\in\mathbb{R}^{(S+1) \times (S+1)}$. Our objective is to transfer the attention vector $\mathcal{R}^l\left[0,1:S+1\right]$, which captures the relationships between the class token and all other tokens. To achieve this, we compute the distance between the teacher's vector and the corresponding vector of the student:
\begin{equation}
D^l = \mathcal{R}_{\mathcal{T}}^l\left[0,1:S+1\right] - \mathcal{R}_{\mathcal{S}}^l\left[0,1:S+1\right]
\label{eq:dist}
\end{equation}
Where the subscripts $\mathcal{T}$ and $\mathcal{S}$ denote the teacher and the student respectively. This approach ensures that the student model maintains the relationships between the class token and the other tokens aligned to those learned by the teacher model.

\paragraph{Adapter networks}
The aforementioned knowledge transfer technique is an effective strategy for alleviating the issue of forgetting the pretraining state. However, it carries the risk of making the model too rigid, hindering its ability to adapt to new tasks. To address this concern, we introduce additional learnable components which we call \textbf{\textit{adapters}}. These adapters aim to prevent the transfer of irrelevant information from the teacher model to the student model. To achieve this objective, we feed the attention vectors $\mathcal{R}_{\mathcal{T}}$ generated by the teacher model into these adapter modules. The adapters are tasked with producing binary masks denoted as $\mathbb{M(\cdot)}$. To obtain the binary masks, we multiply the attention vectors by two sets of learnable weights, resulting in a vector $\textbf{v} \in \mathbb{R}^{(S+1) \times 2}$. We then apply binary Gumbel-Softmax sampling on the last dimension, obtaining a binary vector. Next, we multiply the binary vectors by the distance vector $\mathcal{D}$ derived from equation \ref{eq:dist}. Subsequently, we calculate the mean squared distance and average this value across all the adapters. The objective function for this process can be defined as:
\begin{equation}
    \label{eq:obj_2}
     \mathcal{L}_{FP} = \frac{1}{|L|} \bigg[ \sum_{l \in L} ||\mathbb{M}(\mathcal{R}_{\mathcal{T}}^l) \odot \mathcal{D}^l||_2^2 \bigg],
\end{equation}
Here, $\mathbb{M}^l(\mathcal{R}_{\mathcal{T}}^l)$ represents the binary vector obtained by the $l^{th}$ adapter using the teacher's attention vector $\mathcal{R}_{\mathcal{T}}^l$ and $\odot$ is the element-wise multiplication.

\paragraph{Experience Replay}
Rehearsal methods have proven to be highly effective in addressing the forgetting problem~\cite{farquhar2018towards,aljundi2019gradient}. For this reason, we decided to conduct experiments by incorporating rehearsal techniques into our method. Consequently, at the beginning of the training process, we allocate a fixed-sized memory buffer $\mathcal{M}$. During training, we save a subset of the examples seen by the model into the buffer~\cite{ratcliff1990connectionist,robins1995catastrophic}, along with their associated labels and the logits obtained by feeding these examples to the student network~\cite{buzzega2020dark}. To determine which examples to store in the buffer, we employ the reservoir sampling strategy~\cite{vitter1985random}, which ensures that all examples in the stream have an equal probability of being chosen for storage within the buffer. Moreover, it guarantees that the labels stored in the memory buffer are distributed according to the dataset's label distribution. We want that our model is capable of correctly classifying past examples. This can be achieved by computing the classification loss on the examples sampled from the buffer and using the saved labels as the ground truth:
\begin{equation}
\mathcal{L}_{er} = \mathbb{E}_{(\mathbf{x}_i,\mathbf{y}_i) \sim M} \Big[ \operatorname{BCE}(f_{\theta}(\mathbf{x}_i), \mathbf{y}_i)\Big]
\end{equation}
Another useful strategy for mitigating forgetting on past tasks is to enforce similarity between the student's logits obtained from examples sampled from the buffer and the stored responses. This can be achieved by minimizing the mean squared error between current and saved logits~\cite{buzzega2020dark}:
\begin{equation}
    \mathcal{L}_{der} = \mathbb{E}_{(\mathbf{x}_i,\mathbf{h}_i) \sim M} \Big[ \operatorname{MSE}(f_{\theta}(\mathbf{x}_i), \mathbf{h}_i)\Big]
\end{equation}
Here, $\operatorname{MSE}$ denotes the mean squared error and $\mathbf{h}_i$ represents the logits. By incorporating this additional objective, we encourage the student network to produce logits that closely resemble the stored responses, thereby preserving knowledge learned from previous tasks. 

As an additional step, we apply the ACE technique, as described in \cite{caccia2022new}, which computes the classification loss on the stream examples considering only the labels within the current batch. According to the authors, this approach helps in reducing abrupt changes in model representations.

\paragraph{Adapter mask replay}
As adapter networks contain learnable parameters, they are also susceptible to forgetting. To address this issue, we alleviate the problem by implementing a replay mechanism for the adapter binary masks. Specifically, we store the binary masks generated by the adapters on the non-augmented training samples in a memory buffer. During subsequent tasks, when experience replay is performed and a sample is retrieved from the buffer, we compare the corresponding saved mask with the one currently generated by the adapter networks and compute the following regularization loss:
\begin{equation}
    \label{eq:L_fp_replay}
    \mathcal{L}_{FP\_rep} = \mathop{\mathbb{E}}_{\substack{(\mathbf{x},\mathbf{m}^l) \sim \mathcal{M}\\l \in L}} \bigg[ \operatorname{BCE}\Big(\mathbb{M}(\mathcal{R}_{\mathcal{T}}^l(\mathbf{x})), \mathbf{m}^l \Big) \bigg]
\end{equation}
where $\operatorname{BCE}$ denotes the binary cross entropy loss and $(\mathbf{x},\mathbf{m})$ refers to a pair consisting of a sample and an attention map sampled from the memory buffer.

The final loss function is given by:
\begin{equation}
    \label{eq:obj_2b}
    \mathcal{L} = \mathcal{L}_{clf} + \alpha \cdot \mathcal{L}_{der} + \beta \cdot \mathcal{L}_{er} + \lambda_{FP} \cdot \mathcal{L}_{FP} + \lambda_{FP\_rep} \cdot \mathcal{L}_{FP\_rep}
\end{equation}
where $\alpha$, $\beta$, $\lambda_{FP}$, $\lambda_{FP\_rep}$ are hyperparameters regulating the contributions of the loss terms.

%% file: contents/04_experiments.tex
\subsection{Experimental setup}
\paragraph{Datasets}
We conducted our experiments on two different benchmarks: IIRC CIFAR-100 and WebVision. The former is a recently introduced benchmark, while the latter is a novel benchmark designed by us.

The \textbf{IIRC CIFAR-100} dataset~\cite{abdelsalam2021iirc} is designed such that every sample is assigned two labels - one for the CIFAR-100~\cite{krizhevsky2009learning} superclass and the other for the subclass. There are 22 tasks in total, with the first task comprising only the superclass labels, while the subsequent tasks introduce subclass labels. To make the task more challenging, we opted for the \textbf{incomplete information setting}, where only the label corresponding to the current task is provided for each sample.

The \textbf{WebVision} dataset~\cite{li2017webvision} includes over 2.4 million web images sourced from the Internet, each of which is accompanied by metadata such as a title, description, and tags. Using the provided tags, we create a challenging real-world incremental multi-label benchmark that we refer to as \textbf{Incremental WebVision} by following these steps:
\begin{enumerate}[leftmargin=*]
    \item Starting from a pool of $816,814$ tags, we discard those with less than $1000$ occurrences across the dataset so as not to train the model on targets which are not general enough;
    \item Using feature similarity in CLIP-space~\cite{radford2021learning} as an indicator of semantic proximity, we manually review individual clusters of consistently co-occurring tags and merge those that are semantically overlapped. As a result, we obtain a curated list of roughly $350,000$ images with $300$ tags, that will respectively serve as examples and (multi-)labels for our benchmark;
    \item During each task $T_i$ ($i\in1,..,6$) we first choose and introduce $50$ unique new labels $\mathcal{Y}_i$ and subsequently compose the task's dataset by choosing $\approx58,000$ ($\approx350,000/6$) images such that they are associated with at least one tag in $\mathcal{Y}_{1..i} \triangleq \bigcup_{j=1}^{i} \mathcal{Y}_j$\footnote{Our benchmark adheres to the \textbf{complete information} setup as formulated in the IIRC paper~\cite{abdelsalam2021iirc}.};
    \item Finally, we reserve $5000$ images from each task for testing and employ the remaining $53,000$ for training purposes.
\end{enumerate}
The dataset obtained from the steps above portrays a particularly challenging and real-world scenario, as annotations in WebVision are known to contain significant noise~\cite{li2017webvision}.

\paragraph{Metrics}
As stated in \cite{sorower2010literature}, conventional metrics fail to capture the varying levels of correctness in multi-label predictions. To address this limitation, building on the IIRC framework, we decided to use the \textbf{Precision-Weighted Jaccard Similarity} (\textbf{PWJS}) as our evaluation metric, defined as:
\[
R_{j,k} = \frac{1}{n_k} \sum_{i=1}^{n_k} \frac{|Y_{j,k,i} \cap \hat{Y}_{j,k,i}|}{|Y_{j,k,i} \cup \hat{Y}_{j,k,i}|} \times \frac{|Y_{j,k,i} \cap \hat{Y}_{j,k,i}|}{|\hat{Y}_{j,k,i}|}
\]
Where $R_{j,k}$ is the performance of a model on task $k$ after training on task $j$, $\hat{Y}_{j,k,i}$ are model predictions for the $i^{\text{th}}$ sample of task $k$ after training on task $j$, $Y_{j,k,i}$ are the corresponding ground-truth labels, and $n_k$ is the number of samples of the test set of task $k$.

Since we are dealing with a sequence of tasks, we summarize the final performance by considering the average value of $R_{j,k}$ after training on the last task, which we refer to as the \textbf{Final Average PWJS}:
\begin{equation}
    \label{eq:bam}
    AR_{f} = \frac{1}{N} \sum_{m=1}^{N} R_{N,m}
\end{equation}
where $N$ is the total number of tasks in our benchmark.
We run all our experiments from the same pretrained weights, repeating each 2 times and reporting the average $AR_f$ and standard deviation for all evaluated approaches.

\paragraph{Competitors}
In order to assess the effectiveness of our proposed approach, we conducted a thorough evaluation by comparing it with the following state-of-the-art Continual Learning techniques:
\begin{itemize}[leftmargin=*]
\setlength\itemsep{0em}
    \item \textbf{Experience Replay (ER)}~\cite{ratcliff1990connectionist} is a simple rehearsal strategy which consists in saving the samples seen by the model in a memory of a fixed size. During the continual training, the examples from the memory are given to the model again, alongside the examples of the current task. Although this approach is straightforward, it has the drawback of leaving a significant memory footprint.
    \item \textbf{ER with Asymmetric Cross-Entropy (ER-ACE)}~\cite{caccia2022new} stands out from ER by computing the loss function for examples in the current batch considering only the classes present in that batch. Instead, for examples sampled from the buffer, it calculates the loss function considering all the classes seen up to that point.
    \item \textbf{DER++-ACE}~\cite{caccia2022new} combines the logit distillation-based approach presented in~\cite{buzzega2020dark} with the insights offered by ER-ACE.
    \item \textbf{Learning to Prompt} (\textbf{L2P})~\cite{wang2022learning} creates a pool of learnable prompts each associated to a learnable key. During training, the representation of a sample is obtained from a pretrained and frozen ViT, which is then used to query the prompt pool by computing the similarity between the query and each key. The prompts with the most similar keys are then appended to the input of the blocks of a Vision Transformer, with only the prompts being trained, while the backbone remains frozen.
    \item \textbf{CODA-Prompt (CP)}~\cite{smith2022coda} allocates a set of prompt components for each task. Unlike L2P, this method calculates a weighted sum of the prompts, using the similarity values between the query and keys as weights. Additionally, the approach employs attention vectors with learnable values to filter the queries, thus enabling the model to focus only on their relevant features.
\end{itemize}

Additionally, we contextualize our results by presenting the following additional results: \textbf{Finetuning}, a lower bound where no continual learning technique is applied during training on the task sequence, \textbf{Finetuning with Asymmetric Cross-Entropy (Finetuning-ACE)}, which applies Cross-Entropy selectively on current task data as prescribed in~\cite{caccia2022new} as a baseline passive countermeasure to forgetting, and \textbf{Joint training}, an upper-bound approach which assumes that all the examples are available and the model is trained on the entire dataset without learning incrementally.

\subsection{Results}
\begin{table}[h]
    \caption{Final Average PWJS (and standard deviation) over the task sequence for the IIRC and WebVision benchmarks. The lower part of the table includes methods that require the memory buffer for rehearsal.}
    \input{contents/results_table}
    \label{tab:results}
\end{table}

The performance disparity between the two benchmarks is evident from the results presented in Table~\ref{tab:results}. Across all methods, there is a notable struggle when dealing with the WebVision dataset, which we attribute to the high presence of annotation noise in the dataset. Moreover, on IIRC, prompting-based methods fail to achieve satisfactory performance, with the more sophisticated Coda-Prompt being outperformed by L2P. This is in contrast with traditional single-label benchmarks, which usually depict these advanced methods as best performers. We argue that such an outcome remarks the importance of our work for more complex scenarios.

As we move to replay methods, we observe that, in general, these approaches outperform prompting methods on both benchmarks. Notably, the ACE technique plays a crucial role in enhancing the model's performance. Indeed, this shows the importance in a multi-label setting of confining the contribution of each sample to avoid interference between tasks. Furthermore, the incorporation of logits distillation in DER++ leads to further improvements in the final performance.

Finally, our proposed method shows superior performance across all settings and for all choices of buffer size, which remarks the importance of adapting and maintaining the knowledge from the pre-training.

%% file: contents/results_table.tex
\begin{center}
\setlength{\tabcolsep}{2pt}
\begin{tabular}{l@{\hskip 0.3cm}cc@{\hskip 0.3cm}cc}
\toprule
$\boldsymbol{AR_f}\,\%$ & \multicolumn{2}{c}{\textbf{IIRC-CIFAR}} & \multicolumn{2}{c}{\textbf{Incr.\ WebVision}}\\
\midrule
Joint (UB)        & \multicolumn{2}{c}{\result{86.83}{0.22}}                  & \multicolumn{2}{c}{\result{25.12}{0.01}}      \\
Finetuning-ACE        & \multicolumn{2}{c}{\result{3.21}{0.01}}               & \multicolumn{2}{c}{\result{2.34}{0.36}}     \\
Finetuning (LB)        & \multicolumn{2}{c}{\result{0.24}{0.01}}               & \multicolumn{2}{c}{\result{18.81}{0.59}}     \\
\midrule
L2P \cite{wang2022learning} & \multicolumn{2}{c}{\result{3.76}{0.01}} & \multicolumn{2}{c}{\result{0.31}{0.25}} \\

CP \cite{smith2022coda} & \multicolumn{2}{c}{\result{3.79}{0.01}} & \multicolumn{2}{c}{\result{15.05}{1.45}} \\
\midrule
\textbf{Buffer Size} & 500 & 2000 & 2000 & 5000\\
\midrule

ER \cite{ratcliff1990connectionist,robins1995catastrophic}  & \result{21.04}{1.81} & \result{38.55}{0.91} & \result{17.76}{0.18} & \result{21.63}{0.18} \\
ER-ACE \cite{caccia2022new}  & \result{41.83}{1.75} & \result{50.66}{0.87} & \result{7.75}{1.78} & \result{16.26}{0.34} \\

DER\texttt{++}-ACE \cite{buzzega2020dark}  & \result{41.66}{2.55} & \result{61.82}{0.09} & \result{8.97}{0.01} & \result{17.03}{0.01} \\

\textbf{SCAD (ours)} & \result{\textbf{45.23}}{0.03} & \result{\textbf{66.58}}{0.02} & \result{\textbf{20.02}}{0.35} & \result{\textbf{22.17}}{0.60
} \\

\bottomrule
\end{tabular}
\end{center}

%% file: contents/05_analysis.tex
\subsection{Contribution of selective distillation}
\begin{wraptable}{r}{6cm}
\vspace{-1.2cm}
\setlength{\tabcolsep}{5pt}
\caption{\small Final PWJS achieved by our method on the IIRC CIFAR100 benchmark with and without the use of adapters for generating the binary masks to filter the knowledge transfer.}
\vspace{0.3cm}
\begin{tabular}{l@{\hskip 0.3cm}cc@{\hskip 0.3cm}cc}
\toprule
 & \multicolumn{4}{c}{\textbf{Binary masks}} \\
& \multicolumn{2}{c}{\cmark} & \multicolumn{2}{c}{\xmark} \\
\midrule
Final PWJS        & \multicolumn{2}{c}{\result{66.58}{0.02}}  & \multicolumn{2}{c}{\result{63.68}{0.01}}  \\
\bottomrule
\end{tabular}
\label{tab:selective_distillation}
\vspace{-0.5cm}
\end{wraptable}

To assess whether the inclusion of adapter neural networks has a positive impact on the model's performance, we attempted to remove them from the model and perform distillation using the entire class attention vectors. The results are presented in Table \ref{tab:selective_distillation}. As observed, the filtering of attention vectors performed through the binary masks produced by the adapters improved the final performance. Consequently, it can be inferred that the adapter networks are able to effectively select and propagate only the relevant information from the teacher's attention vector to the student network.

\subsection{Quantitatively measuring catastrophic forgetting in Multi-Label Continual Learning}
\begin{table}[h]
    \caption{Multi-Label Adjusted Forgetting $FG_f$ (the lower the better) of methods evaluated in our benchmarks.}
    \input{contents/forgetting_table}
    \label{tab:forg}
\end{table}

While $AR_f$ provides a clear snapshot of a model's accuracy at the end of training, we are also interested in measuring how the model's performance degrades as the learner is subject to a growing amount of data. For this purpose, we take inspiration from traditional CL evaluation and formulate a MLCL version of the Adjusted Forgetting measure~\cite{bonicelli2023effectiveness,chaudhry2018riemannian} based on PWJS:
\begin{equation}
  \begin{gathered}
    FG_f = \frac{1}{N-1} \sum_{m=1}^{N-1} \bigg[\frac{R_{*,m} - R_{N,m}}{R_{*,m}}\bigg]^+, \\
    \text{where}~~ R_{*,m} = \max_{t\in\{m,\dots,N-1\}}R_{t,m},~~\forall m \in \{1,\dots,N-1\}~~\text{and}~~ [\cdot]^+=max(\cdot, 0).
  \end{gathered}
\end{equation}
$FG_f$ ranges between $0$ and $100$: $FG_f=100$ indicates a result with no retention of performance on previous tasks; $FG_f=0$ denotes no performance decrease on past tasks. 

In Tab.~\ref{tab:forg}, we report the measurements of $FG_f$ for the experiments in Tab.~\ref{tab:results}. We clearly observe that the ACE technique proves highly effective in reducing the forgetting value on IIRC benchmark, indicating that computing the loss based on the classes of the current batch helps to preserve the parameters of the network that encode information from past tasks. On the WebVision benchmark, the situation is different, and for the rehearsal techniques the forgetting values are quite similar: we attribute this to the fact that all methods struggle to achieve good results in this setting. 

Regarding prompting techniques, we note that the two tested methods experience similar levels of forgetting in the IIRC benchmark, while L2P exhibits significantly more forgetting on WebVision. In general, rehearsal-based methods tend to have lower forgetting values. This suggests that in a multi-label setting, having access to a memory of past examples is an effective strategy for retaining past knowledge.

%% file: contents/forgetting_table.tex
\begin{center}
\setlength{\tabcolsep}{2pt}
\begin{tabular}{l@{\hskip 0.3cm}cc@{\hskip 0.3cm}cc}
\toprule
$\boldsymbol{FG_f}$ & \multicolumn{2}{c}{\textbf{IIRC-CIFAR}} & \multicolumn{2}{c}{\textbf{Incr.\ WebVision}}\\
\midrule
Finetuning-ACE        & \multicolumn{2}{c}{65.40}               & \multicolumn{2}{c}{74.45}     \\
Finetuning (LB)        & \multicolumn{2}{c}{98.67}               & \multicolumn{2}{c}{37.39}     \\
\midrule
L2P & \multicolumn{2}{c}{97.29} & \multicolumn{2}{c}{66.41} \\

CP & \multicolumn{2}{c}{97.77} & \multicolumn{2}{c}{44.06} \\
\midrule
\textbf{Buffer Size} & 500 & 2000 & 2000 & 5000\\
\midrule

ER  & 69.98 & 41.57 & 34.33 & 25.64 \\
ER-ACE  & 45.25 & 40.06 & 57.52 & 32.80 \\

DER\texttt{++}-ACE  & 25.82 & 17.71 & 48.95 & 31.87 \\

\textbf{SCAD (ours)} & 24.36 & 16.40 & 28.37 & 24.92 \\

\bottomrule
\end{tabular}
\end{center}

%% file: contents/06_conclusions.tex
In this work, we acknowledge the success of recent CL approaches in contrasting \textit{catastrophic forgetting} in the standard single-label benchmarks. However,  we reveal through a novel and challenging benchmark that state-of-the-art prompting CL methods fail in a multi-label setting, while good old-fashioned rehearsal techniques maintain good performance.

Hence, we present \methnam: a baseline model that tackles the challenges posed by the multi-label setting. Our proposal combines the principles of traditional rehearsal techniques with learning strategies based on knowledge transfer. \methnam exploits the finding that large-scale pretrained architectures are robust against catastrophic forgetting per se; therefore, we have designed a solution that discourages the model from deviating too far from its pretraining state. 

By virtue of a comprehensive experimental analysis, we show that our method achieves superior performance compared to current state-of-the-art approaches in the Multi-Label Continual Learning setting. We recommend \methnam as a strong starting baseline to tackle the challenges that this new era of Continual Learning has to offer.